\documentclass{article}

\usepackage{amsmath}
\usepackage{url}
\usepackage{xspace}
\usepackage[comma,authoryear]{natbib}
\usepackage{todonotes}

\usepackage{mathtools}
\DeclarePairedDelimiter\floor{\lfloor}{\rfloor}
\DeclareMathOperator*{\argmin}{arg\,min}

\usepackage{algorithmic,algorithm}

\newcommand{\myv}{\ensuremath{\textbf{v}}\xspace}

\author{Matthias Gall\'e \\
Xerox Research Centre Europe \\
France}

\title{What Can I Do Now? \\ Guiding Users in a World of Automated Decisions}
\begin{document}

\maketitle

\begin{abstract}
More and more processes governing our lives use in some part an automatic decision step, where -- based on a feature vector derived from an applicant -- an algorithm has the decision power over the final outcome. 
Here we present a simple idea which gives some of the power back to the applicant by providing her with alternatives which would make the decision algorithm decide differently. 
It is based on a formalization reminiscent of methods used for evasion attacks, and consists in enumerating the subspaces where the classifiers decides the desired output. 
This has been implemented for the specific case of decision forests (ensemble methods based on decision trees), mapping the problem to an iterative version of enumerating $k$-cliques.
\end{abstract}

We live in a world where more and more of decision affecting our lives are taken by automatic systems. 
This is of low concern if it affects the advertisements we receive, or the movies/books/products we are recommended to watch/read/buy.
Their use becomes questionable however when it forces us into situation where we can do nothing about, like receiving a sentence, not receiving a credit or having the cost of our medical or car insurance increased.
In particular, a recurrent criticism of autmomatic decision system is that they take away any possibility of human reaction to their decision.

In the words of a very recent best-selling book~\citep{ONeil2016}: 

\begin{quote}

Human beings employing the programs deliver unflinching verdicts, and the humans being that employ them can only shrug, as if to say: "Hey, what can you do"

[..]

Their verdicts, even when wrong or harmful, were beyond dispute or appeal

\end{quote}


This has created a lively debate in the research community and created a field called ``algorithmic fairness''.
High-profile cases, like the use of automatic risk-assessment for criminals\footnote{\url{https://www.propublica.org/article/machine-bias-risk-assessments-in-criminal-sentencing}}, high-paid jobs adverted more to men than women\footnote{\url{https://www.theguardian.com/technology/2015/jul/08/women-less-likely-ads-high-paid-jobs-google-study}} or predictive policing\footnote{\url{http://internetactu.blog.lemonde.fr/2015/06/27/police-predictive-la-prediction-des-banalites/}} also brought this into the general audience.

We argue that the problem of these systems is their removal of any option for the human, and the associated feeling of despair.
It seems naive to think that companies and administration will stop using those systems completely.
Current solutions like open-sourcing the model or the algorithms, adding justification of the decisions or ensuring some notions of fairness are all valid, and should be part of those systems.

In this line, we study in this paper the particular problem of providing the user with a set of steps which she has to do in order to achieve the desired outcome.
Our proposal is a complementary solution, which does not require to disclose details about the model.
The system requires the user to weight the relative cost of changing the features, weight which could be linear, infinite (changing height, or getting younger)
, quadratic (loosing weight) or any other function, not necessarily differentiable nor symmetric.
Based on that, the systems recommends an alternative set of features that minimizes the modification cost but ensures that the decision of the algorithm would change.
Such a tool can now be used either independently by the end-user, or it could be part of a solution provided to an intermediate human agent whose interest it is to provide a positive solution but without raising red flags in his institutional system.

The basic idea is to enumerate all subspaces where the classifier would provide the desired decision, and returning those that are close enough to the original feature vector, with respect to the cost function, which can be very flexible and user-specific.
We show that this approach is feasible for the concrete case of ensemble methods of decision-trees.

\section{Problem formalization}
We will work with binary classifiers, and -- without loss of generality -- will assume that the starting point is a trained classifier $f$ (which we will not modify) and a feature vector \myv which is classified as class $c$ ($0$ or $1$): $f(\myv)=c$.
We are interested in modifying $\myv$ the least possible in order to obtain $\myv'$ that gets classified as the opposite class $c'=1-c$:

\begin{equation}
\displaystyle \myv' = \argmin_{\myv' | f(\myv')=c'} d(\myv,\myv')
\label{eq:thegrial}
\end{equation}

Differently from other work, we do not want to restrict the distance/cost function $d$ to be a norm, but to keep it general, possibly even relaxing metric assumptions.

The only restriction we impose is to define $d$ component-wise, as the sum of the cost of passing from one feature value to another: 

\begin{equation}
\displaystyle  d(\myv,\myv') = \sum_{i=0}^{|\myv|} d(\myv[i],\myv'[i])
\label{eq:cost}
\end{equation}

\noindent where the component-wise cost function is user-specific, as different user may value one attribute more than another.

\section{Related Work}

Many methods for resolving Eq.~\ref{eq:thegrial} have been proposed, for different classifiers, with the final application of \textit{active learning} or, more closely, \textit{evasion attacks}.
The specific solutions varies with the concrete instantiation of the classifier $f$ and the cost function $d$.
If $f$ is differentiable, then a projected gradient descent method can be used~\citep{Biggio2013}, although the projection onto the valid subspace with ensures $d(v,v') < d_\text{max}$ may cause problems with local minimums, or deviating too much from the original gradient.

\citet{Kantchelian2016} addresses the problem of evading  decision tree, relying on integer linear programming techniques.
Both their exact solutions and an heuristics are studied only in the case where the cost to be minimized is a norm ($\ell_p$), treating each feature equally.
This precludes cases where the features are meaningful attributes (instead of, say, pixels), some of which can not be changed, and where the cost may vary greatly.
~\citet{Lowd2005} goes beyond that, weighting each dimension with a feature-specific weight.
However, the cost is still a distance (the $\ell_1$ norm, and therefore symmetric and linearly dependent on the distance).

\cite{Dalvi2004} considers the general case, where no further restrictions are based on $d$ (defined as in Eq.~\ref{eq:cost}), and proposes a learning strategy that takes into account the presence of an adversary using tools from game theory.
The classifier $f$ there is considered to be Na\"ives Bayes,


\section{Algorithm for Tree-Ensembles}
Existing solutions, such as those shown in~\citet{Biggio2013}, can be applied to linear, SVM or neural network-based classifiers.
Here we propose a novel solution for tree-ensembles: these are very efficient non-linear classifiers which are widely used in industrial applications.
We assume the ensemble forest is composed of $k$ binary decisions trees where at each node $n$ a single-feature threshold decision is made, dividing the remaining data-points into two sets, depending weather feature $x^{(n)}$ is smaller or equal; or larger than threshold $\tau_n$ (this does not take into account so-called \textit{oblique}~\citep{Menze2011} trees which however are rarely used in practice).
Each leaf $n$ is associated with an outcome $\textit{class}(n)$, and each tree classifies an entry according to the leaf associated to the sequence of decision in the path from the root.
The ensemble method uses simple voting to determine the final prediction.

Our algorithm consists in mapping leaf nodes of the class $c'$ to nodes of a graph, and adding edges if the sets they respectively restrict overlap.
The subspace for which the ensemble classifier would then predict $c'$ is therefore defined by cliques of size $\floor*{\frac{k}{2}}+1$.

\subsection{Construction of Graph}
We now denote formally the construction of our undirected graph $G=(V,E)$, with $V$ as usual denoting the set of vertices and $E$ the set of pairs denoting edges.
Each leaf node $i$ of class $c'$ of decision tree $j$ will correspond to a node  in the graph $G$:

\[ \displaystyle  V = \left\{n_i^{(j)} \left|  1 \leq j \leq k,  \textit{class}\left(n_i^{(j)}\right) = c', n_i^{(j)} \text{is a leaf node of }t_j \right. \right\} \]

$(n_{i_1}^{(j_1)},n_{i_2}^{(j_2)})$ will be in $E$, if the following conditions hold:

\begin{itemize}
	\item the intersection of their corresponding intervals should be non-empty (which in particular implies that $j_1 \neq j_2$).
	\item they should denote a consistent solution: A consistent solution refers to potential global constraints due to the representation of qualitative attributes in the feature space. 
	For instance, a persons gender may be encoded as an indicator vector, but an interval which forces both to be $0$ is not consistent (given an indicator encoding of length $m$, of the $m$ interval restrictions at least one has to admit a $1$ and $m-1$ have to admit a $0$).
\end{itemize}

With this graph, any clique of size at least $\floor*{\frac{k}{2}}+1$ corresponds now space where the random forest would predict class $c'$ as outcome.
We propose therefore to enumerate those cliques, filter out inconsistent or empty ones, and to measure their distance to the original feature vector $\textbf{v}$.

Most problems around cliques are of exponential time complexity~\citep[GT19]{GareyJohnson}: enumerating cliques is known since longtime to be polynomial in the output (which can be exponential), and with time delay (the time between two consecutive outputs) of $\mathcal{O}(|E||V|)$~\citep{Tsukiyama1977}.
However, as our experiments (Sect.~\ref{sect:experiments}) show, current implementations of efficient algorithms are fast enough to provide enough samples of cliques in order to be of reasonable use in practice.

\section{Experiments}
\label{sect:experiments}
We implemented the above algorithm, and tested it out on the German Credit Data from UCI\footnote{\url{http://archive.ics.uci.edu/ml/datasets/Statlog+(German+Credit+Data)}}~\citep{Lichman2013}.
Each qualitative attribute (13 out of 20) were encoded as indicator vectors, while the other 7 numerical ones were used in their original form.
These features include gender, credit history, savings, employment status, gender and many more.
This is a binary classification problem, where each feature vector is labeled as good or bad.
Random Forest (using 10 decision trees) achieves an accuracy of $74.6\%$ on 3-fold cross-validation, in line with what is reported in the literature~\citep{Ann2002,Odea2001} and similar to other classifiers (form the ones we tried, it outperforms nearest-neighbor, Naive Bayes with various priors, and SVM with various kernels; while only logistic regression obtained a slightly better performance).

For the clique finder we used the Parallel Maximum Clique (PMC) Library\footnote{\url{https://github.com/ryanrossi/pmc}}~\citep{Rossi2013} which proved to be very fast.
Those experiments can be modified interactively at \url{http://empowerhuman.xrce.xerox.com:8000/notebooks/WhatCanIDo.ipynb}.
In particular, the file containing the weights can be modified in order to obtain different results.
Evaluation of such approaches is non-obvious, and we present those experiments as a proof-of-concept that our proposal of enumerating the subspaces of the classifier is feasible.

\section{Conclusions}

We argue in this paper for decision systems that provide help to the user, in the form of concrete actions she can perform in order to obtain a desired output.
The user has to specify a user-specific cost function, which can be of any form, allowing comparison between different changes of feature values.
Our method consist in enumerating all the subspaces where the classifier would provide the desired decision.
In the case of forests of decision trees, enumerating these subspaces can be mapped to the problem of enumerating $k$-cliques, for which efficient implementation exists.

\bibliographystyle{authordate1}
\bibliography{whatcanido}

\end{document}